\documentclass[a4paper]{jpconf}
\usepackage{graphicx}
\usepackage{color}
\usepackage{amsmath}
\usepackage{multirow}
\usepackage{amssymb}
\usepackage{pifont}

\newcommand{\cmark}{\ding{51}}%
\newcommand{\xmark}{\ding{55}}%

\begin{document}

\title{Visual Question Answering for Cultural Heritage}

\author{Pietro Bongini, Federico Becattini, Andrew D. Bagdanov, Alberto Del Bimbo}

\address{Media Integration and Communication Center (MICC) - University of Florence.\\ Viale Morgagni 65, Florence, Italy.}

\ead{p.bongini@unifi.it, federico.becattini@unifi, andrew.bagdanov@unifi.it, alberto.delbimbo@unifi.it}

\begin{abstract}
Technology and the fruition of cultural heritage are becoming increasingly more entwined, especially with the advent of smart audio guides, virtual and augmented reality, and interactive installations. Machine learning and computer vision are important components of this ongoing integration, enabling new interaction modalities between user and museum.
Nonetheless, the most frequent way of interacting with paintings and statues still remains taking pictures. Yet images alone can only convey the aesthetics of the artwork, lacking is information which is often required to fully understand and appreciate it. Usually this additional knowledge comes both from the artwork itself (and therefore the image depicting it) and from an external source of knowledge, such as an information sheet. While the former can be inferred by computer vision algorithms, the latter needs more structured data to pair visual content with relevant information.
Regardless of its source, this information still must be be effectively transmitted to the user. A popular emerging trend in computer vision is Visual Question Answering (VQA), in which users can interact with a neural network by posing questions in natural language and receiving answers about the visual content. We believe that this will be the evolution of smart audio guides for museum visits and simple image browsing on personal smartphones. This will turn the classic audio guide into a smart personal instructor with which the visitor can interact by asking for explanations focused on specific interests. The advantages are twofold: on the one hand the cognitive burden of the visitor will decrease, limiting the flow of information to what the user actually wants to hear; and on the other hand it proposes the most natural way of interacting with a guide, favoring engagement.
\end{abstract}

\section{Introduction}
Museum visits have adapted throughout the years to exploit technological advances. Nowadays cultural heritage heavily relies on some form of multimedia content to deliver information to the user in ways that limit cognitive burden and engage the visitor as much as possible. This is especially true for young visitors, where gamification techniques have often proven effective~\cite{becattini2016imaging, ioannides2017mixed}. Technology can help bridge the gap between user interests and the message the museum wants to convey.

Videos, 3D reconstructions and augmented realities, among others, have become an integral part of the visit, which has now shifted its focus not solely on artworks but also on how they are organized and presented. To offer a richer experience, smart audio guides have also been developed, gradually replacing information sheets or offering some sort of augmented visit relying on sensors available on personal smartphones.
Despite the increasing diffusion of devices to help guide the visitor, the most effective way to convey most information still remains a human guide with whom the visitor may interact to ask for clarifications or deeper discussions on topics of interest. In fact, the user requires a natural way to interact with whomever is providing the information, be it an actual museum guide or a piece of software.

At the same time, the diffusion of personal assistants on smartphones is aiding an increasing number of people with everyday tasks. These assistants, though, still offer little or no help in the area of cultural heritage. This is due to the need to process complex pieces of structured information, which are often transversal to several domains.

Machine Learning is starting to reach out to the complexity of these tasks. In particular the emerging topic of \emph{Visual Question Answering} is able to engage a user by answering questions about visual media ~\cite{antol2015vqa, goyal2017making}. VQA algorithms merge the capabilities of Computer Vision to understand image content and those of Natural Language Processing to reason about questions and provide relevant answers. VQA builds upon the Question Answering literature, where questions are answered related to text instead of visual content.
Interest in VQA has grown quickly, but it has still not been applied to cultural heritage since the knowledge required to answer the variety of questions a user might ask about an artwork are not contained within the opera itself. A full understanding requires external knowledge usually obtainable only from experts (e.g. museum guides) or information sheets.
This knowledge can be processed separately since it is often available in a textual form, whether it is provided directly from the museum or retrieved from online resources.
Therefore, to be able to address the dual nature of the task, i.e. answering to both visual and contextual questions, VQA and QA must be combined.

In this work we make a first step towards the development of a Visual Question Answering model for cultural heritage by combining the capabilities of a VQA model and a QA model.
Our first contribution is to introduce a module that accurately discriminates between visual and contextual questions. Our second contribution is to design a model made of two branches able to answer to both kinds of questions. Our experiments demonstrate the effectiveness of our technique for question classification the performance of our general question answering model.
To evaluate our model, we annotated a subset of ArtPedia \cite{stefanini2019artpedia} with visual and contextual question-answer pairs.\footnote{Our dataset will be publicly released upon publication of this work.} 

In the next section we briefly review works from the literature relevant to our contribution. In section~\ref{sec:method} we describe our approach to integrating Visual and Contextual Question Answering and Contextual for the cultural heritage domain, and in section~\ref{sec:experiments} we report on a number of experiments we performed to quantify the performance of our approach. We conclude in section~\ref{sec:conclusions} with a discussion of our contribution. 

\section{Related Work}
\label{sec:related}

Visual Question Answering (VQA) is an emerging topic which aims at automatically answering open-ended questions about a specific image. Together with image captioning~\cite{liu2017improved}, VQA is the main point of contact between the communities of Natural Language Processing (NLP) and Computer Vision.
This intersection consists in a great variety of sub-tasks like question reasoning~\cite{lu2016hierarchical}, object detection~\cite{han2018advanced}, object recognition~\cite{kheradpisheh2018stdp}, etc.
For this reason, VQA requires a high-level understanding of images and questions. A limitation of Visual Question Answering techniques is that they are not able to answer questions whose answers require  knowledge external to the image (e.g. Who is the man in the image? How is this person called?). These kind of questions are arguably the most interesting since humans are likely to ask questions about what they are not able to deduce from the image.

Most approaches in VQA are based on Deep Learning and use Convolutional Neural Networks (CNNs)~\cite{donahue2013decaf} to interpret images and Recurrent Neural Networks (RNNs)~\cite{mao2014deep} to interpret sentences or phrases. The extracted visual and textual feature vectors are then typically jointly embedded by concatenation, element-wise sum, or product to then infer an answer. In~\cite{anderson2018bottom} Anderson \textit{et al.} designed a bottom-up attention mechanism based on salient objects in the images. In particular, instead of considering the entire image divided in cells (as done in previous methods~\cite{shih2016look}) they use object features as attention candidates. These features are extracted using a detector such as Faster R-CNN \cite{ren2015faster} trained on the Visual Genome dataset~\cite{krishnavisualgenome}. This technique was an important step forward for the VQA community and increased VQA performance considerably. In~\cite{cadene2019murel, yu2019deep} the authors use the bottom-up objects candidates together with the modules that they developed to achieve state-of-the-art performance.

\section{Method}
\label{sec:method}

In this section we describe our approach to open-ended visual question answering. We first show the general model that characterizes our technique then we describe the sub-modules.

\subsection{Visual Question Answering with visual and contextual questions}
The main idea of this work is to classify the type (visual or contextual) of the input question so that the question can be answered by the most suitable sub-model. We rely on a question classifier to understand if the question concerns exclusively visual traits of an image or if an external source of information is needed to provide a correct answer. The question is then fed to a VQA or a QA model, depending on the output of the classifier. In both cases the question must be analyzed and understood, yet the usage for two separate architectures is driven by the need to process different additional sources of information. If the question is visual, then the answer is generated from the image, whereas if the question is contextual then the answer is generated using external textual descriptions.

\begin{figure}
	\centering
	\includegraphics[width=0.7\linewidth]{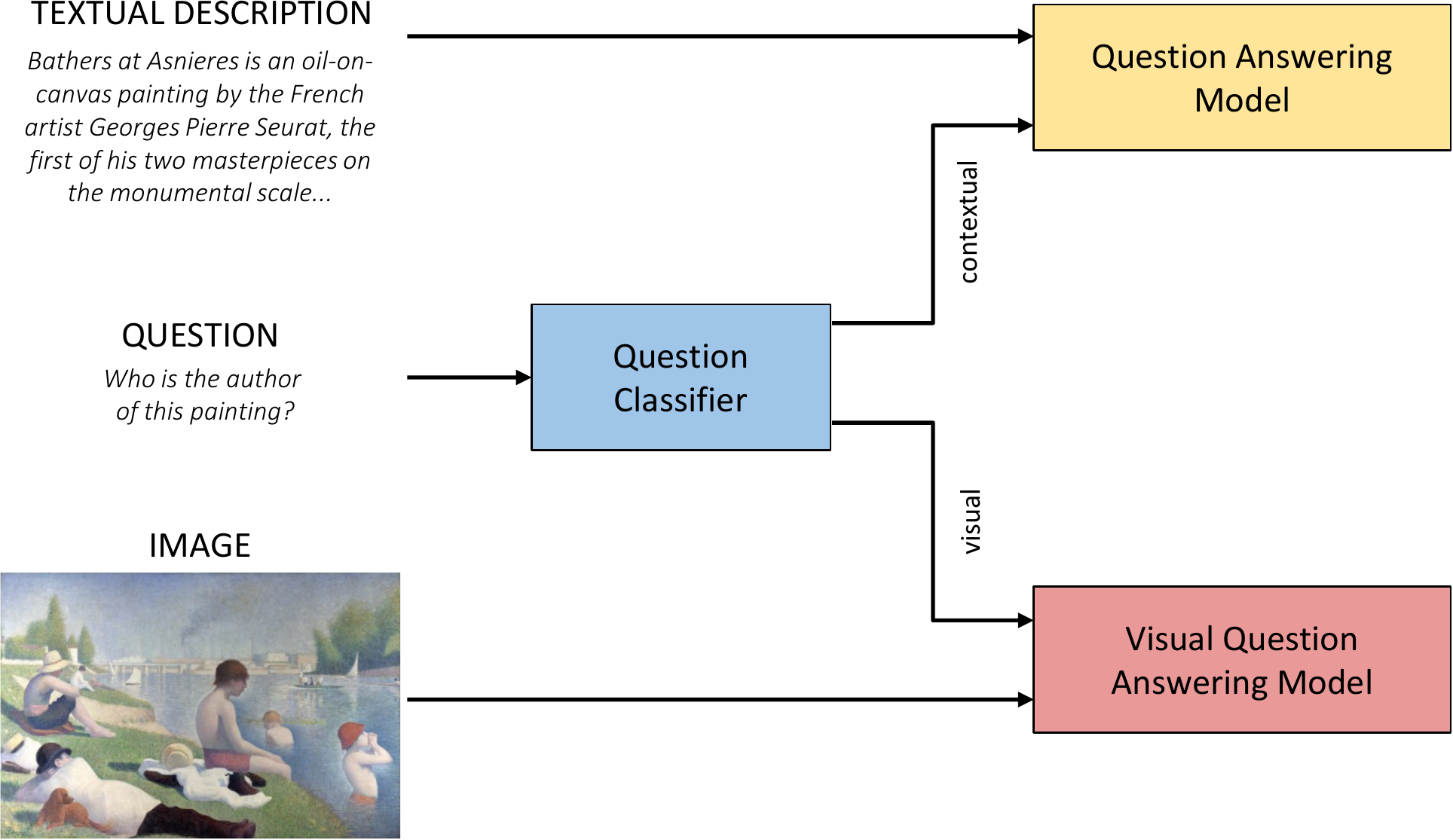}
	\caption{Model overview. A question classifier categorizes the question as visual or contextual. The correspondent module is used to answer the question relying either on the image or external descriptions.}
	\label{fig:model}
\end{figure}

The overall pipeline (see figure \ref{fig:model}) used by our approach to answer a question is the following:
\begin{enumerate}
\item \textbf{Question Classification}. The question is given in input to the question classifier module that determines if the question is contextual or visual.
\item \textbf{[Visual] Question Answering}. Depending on the predicted question type, the corresponding module is activated to generate the answer.
	\begin{enumerate}
	\item If the question is contextual, the question is given in input to a Question Answering module that takes in input also an external information useful to answer the question. This system produces an output answer only based on this external information.
	\item If the question is visual, the question and the image are given as input to a Visual Question Answering module. This system produces an output answer based on the content of the image.
	\end{enumerate}
\end{enumerate}

\subsection{Question Classifier Module}
The question classifier module consists of a Bert~\cite{devlin2018bert} module for text classification. BERT makes use of a Transformer~\cite{vaswani2017attention}, an attention mechanism that learns contextual relations between words (or sub-words) in a text. The Transformer is trained in a bidirectional way in order to have a deeper knowledge of language context and flow. This language model is extremely versatile since it can be used for different tasks like text classification, next word in sentence prediction, question answering and entity recognition. This model is turned into a question classification architecture by adding a classification layer on top of the Transformer output. The input question is represented as the sum of three different embeddings: the token embeddings, the segmentation embeddings and the position embeddings. Moreover, two special tokens are added at the start and in the end of the question.

\subsection{Contextual Question Answering Module}
The Model used for the Question Answering task is another Bert module that focuses on this task. In this case the module takes in input both a question and a textual description. Since this system uses the textual information to answer the question, the text must contain relevant information to generate an appropriate answer.

\subsection{Visual Question Answering Module}
The architecture of the Visual Question Answering module is similar to the one used by Anderson et al.~\cite{anderson2018bottom} in their Bottom-up Top-Down approach. Here the salient regions of the image are extracted by a Faster R-CNN ~\cite{ren2015faster} pre-trained on the Visual Genome dataset~\cite{krishnavisualgenome}. The words of the question are represented with a Glove embedding~\cite{pennington2014glove} and then the question is encoded by a Gated Recurrent Unit (GRU) to condense each question into a fixed size descriptor. An attention mechanism between the encoded question and the salient image regions is developed to weigh the candidate regions that are useful to answer the question. Then the weighted region representations and the question representation are projected into a common space and are joined with an element-wise product. Finally the joint representation passes two fully connected layers and a softmax activation that produces the output answer.

\begin{figure}
	\centering
	\includegraphics[width=0.9\linewidth]{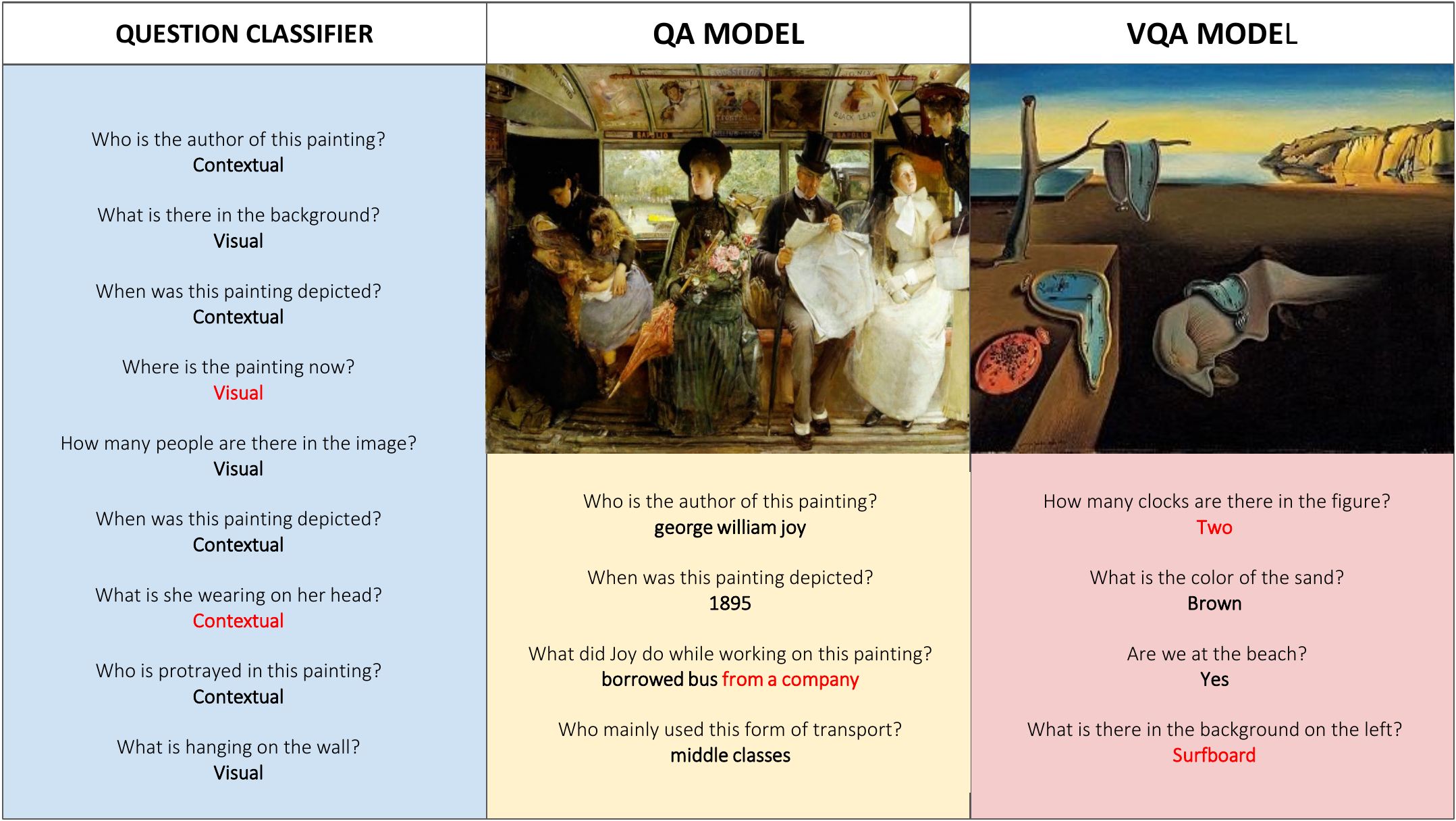}
	\caption{Sample outputs of the three components of our architecture. Correct answers are shown in black, wrong answers in red.}
	\label{fig:qualitative}
\end{figure}

\section{Experimental results}
\label{sec:experiments}

In this section we describe experiments conducted to evaluate the performance of our approach. We first introduce the datasets used for training and testing our network, then we describe the protocols adopted for the experiments and the obtained results.

\subsection{Datasets}
For our experiments we used the standard VQA v2~\cite{balanced_vqa_v2} dataset, OK-VQA~\cite{marino2019ok} and Artpedia~\cite{stefanini2019artpedia}, a dataset containing images of famous paintings.
\paragraph{\textbf{VQA v2}} This dataset contains 443,757 training questions/answers referred to 82,783 training images. The number of test examples is about the same of the training examples, instead the validation examples are about the half. Each image has more questions referred to it and these are of multiple types like relation between objects, activity recognition, counting, object detection and so on. Each question is answered by ten annotators and the given answers compose the ground truth. VQA v2 is currently the most used benchmark for Visual Question Answering tasks.

\paragraph{\textbf{OK-VQA}} OK-VQA is a subset of the VQA v2 dataset and it contains 14,055 open-ended questions where each of these has five ground truth answers. In particular OK-VQA contains all the questions of VQA v2 that cannot be answered with processing only the corresponding image but require external knowledge. We use OK-VQA jointly with the original VQA dataset to obtain sets of questions related to the image (visual questions) or to external knowledge (contextual questions).

\paragraph{\textbf{Artpedia}} The Artpedia dataset contains a collection of 2,930 paintings, each associated to a variable number of textual descriptions collected from WikiPedia. Each sentence is labelled either as a visual sentence or as a contextual sentence, if it does not describe the visual content of the artwork. Contextual sentences can describe the historical context of the artwork, its author, the artistic influence or the place where the painting is exhibited. The dataset contains a total of 28,212 sentences, 9,173 labelled as visual sentences and the remaining 19,039 as contextual sentences. This is not a Visual Question Answering dataset, so we manually annotated a subset of images with both visual and contextual question-answer pairs, based on the available images and descriptions.

\subsection{Experimental protocols}
Our model is composed by three sub-modules: the question classifier that classifies if a question requires visual or contextual information, the question answering module which answers to contextual questions and the visual question answering module which answers to visual questions. The three modules generate different outputs and we evaluate each one of them independently. The Visual Question Answering module answers with short sentences of at most three words chosen from the set of answers. For this reason, as common practice in the VQA literature, we can consider the problem as a classification task and estimate the accuracy to asses its performance:
\begin{equation}\label{eq:1}
 Accuracy = \frac{N_c}{N_a}
\end{equation}{}
where $N_c$ is the number of correct answers and $N_a$ the number of total answers.
The same metric can be used for the question classifier module, since it solves a binary classification task.

The question answering model instead, since it can potentially rely on structured and more complex information from the meta-data, is able to answer to questions with more words, articulating short sentences. For this reason we evaluate its performance not only with Accuracy but also with F1-measure, a metric that takes into account the global correctness of the answer:
\begin{equation}
    F1 = 2 \times \frac{Precision \times Recall}{Precision + Recall}
\end{equation}{}
where $Precision$ is defined as the number of correct words divided by the length of the answer and $Recall$ as the number of correct words divided by the length of the ground truth.

\begin{table}[]
	\centering
	\begin{tabular}{c|c|c}
		& OK-VQA/VQA v2	& Artpedia \\ \hline
		Question Classifier & 0.868 & 0.938 \\
	\end{tabular}
	\caption{\textbf{Question classifier:} accuracy of our question classifier on questions from both the OK-VQA and VQA v2 datasets and from Artpedia.}
	\label{table_question_classifier}
\end{table}

\begin{table}[]
	\centering
\begin{tabular}{ cc }   
\textbf{QA model} & \textbf{VQA model} \\
\begin{tabular}{c|c||c|c}
		Contextual	& Visual	& Accuracy	& F1-score \\ \hline
		\cmark 		& \xmark	& 0.684		& 0.832    \\
		\xmark 		& \cmark	& 0.176		& 0.150    \\
		\cmark 		& \cmark	& 0.504		& 0.417   
	\end{tabular} &  
\begin{tabular}{c|c||c|c}
		Contextual	& Visual	& Accuracy  \\ \hline
		\cmark 		& \xmark	& 0.000	    \\
		\xmark 		& \cmark	& 0.524	    \\
		\cmark 		& \cmark	& 0.251	       
	\end{tabular} \\
\end{tabular}
	\caption{Results of the two answering models on contextual questions, visual questions and both visual and contextual questions from Artpedia. Note that the VQA model does not have access to the external information required to answer the contextual questions, making it unable to answer correctly. See section~\ref{sec:full} for analysis of the performance of our full model on combined Visual/Contextual Question Answering}
	\label{table_vqa_qa_results}
\end{table}




\subsection{Experimental results}
In order to evaluate the performance of our model we make different experiments. We measure the performance of the model analyzing each component independently.

\subsubsection{Question Classifier}
We train the question classifier module with questions of both the OK-VQA and VQA v2 datasets. We take from VQA v2 a number of visual questions equal to the number of questions that require external knowledge from OK-VQA. The obtained dataset is then split into train and test sets.
The question classifier is supposed to understand from the structure of the question whether the answer concerns the visual content or not. This is a generic classifier, agnostic from the domain of the task. In fact, VQA v2 and OK-VQA contain generic images, while we are interested in applications in the cultural heritage domain. We demonstrate the effectiveness of our approach and its ability to transfer to the cultural heritage domain by evaluating it both on the VQA/OK-VQA dataset and on a new dataset comprised of a subset of Artpedia \cite{stefanini2019artpedia}. Since this dataset does not contain questions but only images and descriptions, we took 30 images from this dataset and annotated them with a variable number of both visual and contextual questions (from 3 to 5 for both categories).
The accuracy of our question classifier module is shown in Tab. \ref{table_question_classifier}. We can observe that it is able to predict the type of the question correctly in most cases.

\subsubsection{Contextual Question Answering}
We test our question answering module on the subset of Artpedia containing 30 images that we annotated. In particular, we test the accuracy of our module in three different experiments: test on contextual questions, test on visual questions and test with both visual and contextual questions. Note that the outputs of the visual and contextual modules are different, since VQA is treated as a classification problem, while for QA 
From the results shown in Tab. \ref{table_vqa_qa_results} we can deduce that our question answering module works very well with contextual questions and obtains worse results with visual questions. This can be justified from the fact that visual questions refer to visible details of paintings that cannot be described in visual sentences of ArtPedia.

\subsubsection{Visual Question Answering}
Similarly to the tests conducted for the question answering module, we evaluate the visual question answering module on both visual and contextual questions. In Table \ref{table_vqa_qa_results} results of our visual question answering model are shown. We can observe that conversely from the question answering module this model performs well on visual questions and is not able to answer correctly to contextual questions. This is motivated by the fact that contextual questions require external knowledge (e.g. author, year) that a purely visual question answering engine does not have access to.

\subsubsection{Full pipeline}
\label{sec:full}
Finally, we combine the capabilities of all the modules together and we test on both visual and contextual questions, obtaining an accuracy of 0.570.
The full pipeline, thanks to the question classifier, is able to correctly distinguish between visual and contextual questions. The visual question answering module and the question answering module receive as input almost all questions that they are able to answer (contextual question for the question answering module and visual questions for visual question answering module). For this reason the complete model exceeds the performances of both single answering modules.
Fig. \ref{fig:qualitative} shows some qualitative result of the three components of the pipeline. The components correctly handle most of the questions but some common failure cases can be observed. For instance the Question Answering model might add details to the answer that are not present in the ground truth and the Visual Question Answering model might confuse some elements of the painting with similar objects.

\section{Conclusions}
\label{sec:conclusions}

In this paper we presented an approach for Visual Question Answering in the Cultural Heritage domain. We have addressed two important issues: the need to process both both image and contextual knowledge contained and the lack of data availability. The model we presented combines the capabilities of a VQA and a QA model, relying on a question classifier to predict whether it refers to visual or contextual content. To assess the effectiveness of our model we annotated a subset of the ArtPedia dataset with visual and contextual question-answer pairs.

\section*{Acknowledgements} We gratefully acknowledge the support of NVIDIA Corporation with the donation of the Titan Xp GPU used for this research. This work was partially supported by the project ARS01\_00421: “PON IDEHA - Innovazioni per l’elaborazione dei dati nel settore del Patrimonio Culturale.”

\section*{References}
\bibliographystyle{abbrv}
\bibliography{JPCSLaTeXGuidelines}

\end{document}